%% file: NAC.tex
\documentclass[10pt,twocolumn,letterpaper]{article}

\usepackage{cvpr}
\usepackage{times}
\usepackage{epsfig}
\usepackage{graphicx}
\usepackage{amsmath}
\usepackage{amssymb}
\usepackage{enumerate,booktabs}
\usepackage{subfigure,color}
\usepackage[numbers,sort&compress]{natbib}

\usepackage[pagebackref=true,breaklinks=true,letterpaper=true,colorlinks,bookmarks=false]{hyperref}

\cvprfinalcopy % *** Uncomment this line for the final submission

\def\cvprPaperID{2156} % *** Enter the CVPR Paper ID here

\ifcvprfinal\fi
\begin{document}
\input{definitions}

%%%%%%%%% TITLE
\title{Improving Human Action Recognition by Non-action Classification}

\author{Yang Wang and Minh Hoai\\
Stony Brook University, Stony Brook, NY 11794, USA\\
{\tt\small \{wang33, minhhoai\}@cs.stonybrook.edu}
}

\maketitle
\thispagestyle{empty}

%%%%%%%%% ABSTRACT
\begin{abstract}

In this paper we consider the task of recognizing human actions in realistic video where human actions are dominated by irrelevant factors. We first study the benefits of removing non-action video segments, which are the ones that do not portray any human action. We then learn a non-action classifier and use it to down-weight irrelevant video segments. The non-action classifier is trained using ActionThread, a dataset with shot-level annotation for the occurrence or absence of a human action. The non-action classifier can be used to identify non-action shots with high precision and subsequently used to improve the performance of action recognition systems. 
\end{abstract}

%%%%%%%%% BODY TEXT

\section{Introduction}

\begin{figure}[t]
\begin{center}
\includegraphics[width=1\linewidth]{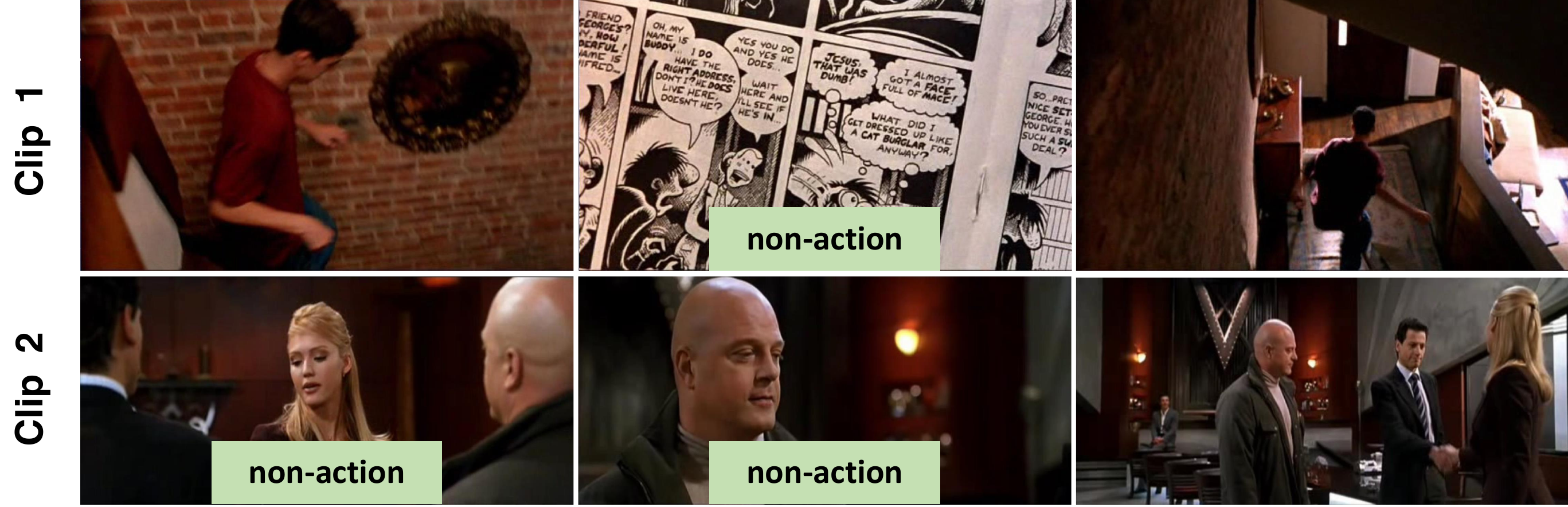}
\end{center}
\vskip -0.1in
   \caption{Examples of non-action shots in typical video clips of human actions. This shows two video clips from the Hollywood2 dataset. Clip 1: the second shot contains no human; Clip 2: the first two shots depict a dialog and exhibit little motion.}
\label{fig:actionEx}
\end{figure}

The ability to recognize human actions in video has many potential applications in a wide range of fields, ranging from entertainment and robotics to security and health-care. However, human action recognition \cite{Aggarwal-Ryoo-ACMCS11,Wang-Schmid-ICCV13,Hoai-Zisserman-ACCV14b, Simonyan-Zisserman-NIPS14,fernando2015modeling,Hoai-et-al-CVPR11,Hoai-Zisserman-CVPR14} is tremendously challenging for computers due to the complexity of video data and the subtlety of human actions. Most current recognition systems flounder on the inability to separate human actions from the irrelevant factors that usually dominate subtle human actions. 
This is particularly problematic for human action recognition in TV material, where a single human action may be dispersedly portrayed in a video clip that also contains video shots for setting the scene and advancing dialog. For example, consider the video clips from the Hollywood2 dataset~\cite{Marszalek-et-al-CVPR09} depicted in Figure \ref{fig:actionEx}. These video clips are considered as `clean' examples for their portrayed actions, but 3 out of 6 shots do not depict the actions of interest at all.
Although recognizing human actions in TV material is an important and active area of research, existing approaches often assume the contiguity of human action in video clip and ignore the existence of irrelevant video shots.

In this paper, we first present out findings on the benefits of having purified action clips where irrelevant video shots are removed. We will then propose a dataset and a method to learn a non-action classifier, one that can be used to remove or down-weight the contribution of video segments that are unlikely to depict a human action. 

Of course, identifying all non-action video segments is an ill-posed problem. First, there is no definition for what a general human action is. Second, even for the classes of actions that are commonly considered such as hug and handshake, the temporal extent of an action is highly ambiguous. For example, when is the precise moment of a hug? When two people start opening their arms or when the two bodies are in contact? Because of the ambiguities in human actions, our aim in this paper is to identify video segments that are unlikely related to the actions of our interest. Many of those segments can be unarguably identified, e.g., video shots that contain no people, show the close-up face of a character, exhibit little motion, or are a part of a dialog (some examples are shown in Figure~\ref{fig:actionEx}). However, instead of manually defining what a non-action segment should be, in this paper we will use supervised learning and train a non-action classifier using video data that has shot-level annotation. The classifier is based on Support Vector Machines~\cite{Vapnik-98} and appearance and motion features. More specifically, we will combine Fisher Vector encoding~\cite{Perronnin-et-al-ECCV10} of Dense Trajectory Descriptors~\cite{Wang-Schmid-ICCV13} and the deep learning features of a Two-stream ConvNet~\cite{Simonyan-Zisserman-NIPS14}.

It should be noted that we propose to learn a non-action classifier for generic human actions. This has several benefits over an action-specific classifier that only aims to identify video segments that are irrelevant to a specific action. First, a generic non-action classifier is universal; it can be used to improve the recognition performance of action classes that do not have detailed training annotation. Second, even when detailed annotation exists, it would still be difficult to obtain a good  action-specific segment classifier. To some extent, having a good classifier that can remove segments that are not related to a specific class is equivalent to having a good action recognizer already. Thus, an action-specific classifier brings no complementary benefits, while a generic action classifier does and can be used to increase the signal-to-noise ratio of actions in video clips.

The rest of this paper is structured as follows. We first review some related topics in Section \ref{sec:relatedwork}. Section \ref{sec:prune_benefit} presents the empirical evidence that pruning irrelevant shots leads to significant improvement on human action recognition. This is followed by the experiment on learning and evaluating a non-action classifier in Section \ref{sec:nonact_clf}. In Section \ref{sec:act_clf}, we propose an approach for using the non-action classifier for human action recognition and describe the performance gains in several experiments.

\section{Related Works} \label{sec:relatedwork}

In this work, we propose to learn a non-action classifier to predict whether a video subsequence is an action instance~\cite{lai2014video}. This is related to defining and measuring objectness in image windows~\cite{Alexe-et-al-CVPR10,alexe2012measuring,uijlings2013selective}, which can assist some common visual tasks like object proposal and detection. Learning such high-level concept often relies on well-defined visual features such as saliency\cite{hou2007saliency}, color, edges~\cite{zitnick2014edge} and super-pixels. There have been some recent attempts~\cite{chen2014actionness,feichtenhofer2015dynamically} to extend objectness to actionness. They often measure the actionness by fusing different feature channels such as space-time saliency \cite{ni2015motion}, optical flow \cite{feichtenhofer2015dynamically}, body configuration~\cite{Hoai-et-al-BMVC14} and deep learning features~\cite{gkioxari2014finding}, sometimes with human input like eye fixation \cite{nguyen2015stap}. However, compared to objectness, actionness in videos is still not sufficiently explored due to the computational intensity in video space and the subtlety of human actions.

\section{Benefits of pruning irrelevant shots} \label{sec:prune_benefit}

We now present our findings on the statistics of non-action shots in a typical human action dataset and the benefits of removing them for human action recognition. We will defer the description of a method for classifying non-action shots to the next section. In this section, we assume there is an oracle for identifying non-action shots.

\subsection{ActionThread dataset}

For the studies in this section, we consider the ActionThread dataset~\citep{Hoai-Zisserman-ACCV14}. This is a typical human action dataset of which the method to collect human action samples is similar to that of many human action datasets, including Hollywood2~\citep{Marszalek-et-al-CVPR09}, TVHID~\cite{Patron-Perez2010}, and Hollywood3D~\cite{Hadfield-Bowden-CVPR13}. The ActionThread dataset consists of video samples for 13 actions, a superset of the actions considered in Hollywood2~\cite{Marszalek-et-al-CVPR09} and TVHID~\cite{Patron-Perez2010}. The video samples were automatically located and extracted around human action occurrences using script mining in 15 different TV series. They are split into training and test sets such that the two subsets do not share samples from the same TV series. 

Amazon Mechanical Turk (AMT) workers were asked to annotate the occurrence of human actions shot-by-shot for each video. Each video shot of the dataset was labeled by three AMT workers. Most (86.3\%) of the shots received the same annotation by three AMT workers. We then manually reviewed and carefully relabeled those shots with conflicting annotations. Of those shots we relabeled, around 60\% were consistent with the majority vote. Videos without action occurrences were eliminated. Finally, we have a dataset of 3,035 videos for 13 actions. Table~\ref{tab:ActionThread} shows detailed statistics for the refined ActionThread dataset. On average, one video contains roughly 6.5 shots, 60\% of which are non-action shots.

\begin{table}[t]
\begin{center}
\begin{tabular}{lrrr}
\toprule
Action & \#Video & \#Shot & \#Non-Act Shot (\%) \\
\midrule
Ans.Phone   & 193 & 1101 & 560  ~(50.9\%) \\
DriveCar    & 68  & 582  & 367  ~(63.1\%) \\
Eat         & 210 & 1185 & 611  ~(51.6\%) \\
Fight       & 136 & 1539 & 582  ~(37.8\%) \\
GetOutCar   & 87  & 620  & 446  ~(71.9\%) \\
ShakeHand   & 113 & 621  & 370  ~(59.6\%) \\
Hug         & 315 & 1745 & 914  ~(52.4\%) \\
Kiss        & 593 & 3365 & 2010 ~(59.7\%) \\
Run         & 738 & 5697 & 3551 ~(62.3\%) \\
SitDown     & 305 & 1560 & 1049 ~(67.2\%) \\
SitUp       & 66  & 386  & 262  ~(67.9\%) \\
StandUp     & 176 & 1048 & 794  ~(75.8\%) \\
HighFive    & 35  & 168  & 117  ~(69.6\%) \\
\midrule
All        & 3035 & 19617 & 11633 ~(59.3\%) \\
\bottomrule
\end{tabular}
\end{center}
\vskip -0.1in
\caption{{\bf Video counts  and the percentage of non-action shots in the ActionThread dataset}. Non-action shots outnumber action shots in most action categories.}
\label{tab:ActionThread}
\end{table}

\begin{table}[t]
\begin{center}
\begin{tabular}{lrrr}
\toprule
&No pruning  & Pruning & Improvement  \\
\midrule
Ans.Phone   & 29.2 & 42.0 & 12.8 \\
DriveCar    & 41.4 & 75.5 & 34.1 \\
Eat         & 33.2 & 56.1 & 22.9 \\
Fight       & 63.0 & 70.6 &  7.6 \\
GetOutCar   & 23.8 & 35.8 & 12.0 \\
ShakeHand   & 42.9 & 58.4 & 15.5 \\
Hug         & 50.1 & 59.4 &  9.3 \\
Kiss        & 65.0 & 72.2 &  7.2 \\
Run         & 85.2 & 93.7 &  8.5 \\
SitDown     & 60.9 & 76.5 & 15.6 \\
SitUp       &  9.5 & 13.2 &  3.7 \\
StandUp     & 40.4 & 58.5 & 18.1 \\
HighFive    & 44.7 & 55.1 & 10.4 \\
\midrule
Mean        & 45.3 & 59.0 & 13.7 \\
\bottomrule
\end{tabular}
\end{center}
\vskip -0.1in
\caption{{\bf Benefits of pruning irrelevant shots.} This shows the Average Precision (AP) of a popular action recognition method, with and without pruning the non-action shots. The last column displays the performance gain for pruning non-action shots. The performance gain is significant, as large as 34\% for DriveCar.}
\label{tab:pruningBenefit}
\end{table}

\subsection{Action recognition system} \label{subsec:rec_system}

We hypothesize that removing non-action shots will improve the recognition performance. We study this hypothesis with a popular action recognition method that holds the state-of-the-art performance on many datasets. This method is based on Dense Trajectory Descriptors~\cite{Wang-Schmid-ICCV13}, Fisher Vector encoding~\cite{Perronnin-et-al-ECCV10}, and Least-Squares Support Vector Machines~\cite{Suykens-Vamdewalle-NPL99}. 

\subsubsection{Trajectory features} \label{subsec:DTD}

The feature representation is based on improved Dense-Trajectory Descriptors (DTDs) \cite{Wang-Schmid-ICCV13}. DTD extracts dense trajectories and encodes  gradient and motion cues  along trajectories. Each trajectory leads to four feature vectors: Trajectory, HOG, HOF, and MBH, which have dimensions of 30, 96, 108, and 192 respectively. 
The procedure for extracting DTDs is the same as \cite{Hoai-Zisserman-ACCV14b,Hoai-Zisserman-ACCV14}, and we refer the reader to \cite{Wang-Schmid-ICCV13,Hoai-Zisserman-ACCV14b,Hoai-Zisserman-ACCV14} for more details.

Note that each trajectory has a temporal span of 15 frames, and the temporal location of each trajectory is taken as the index of the middle frame (the $8^{th}$ frame). For efficiency, we only extract trajectory descriptors for each video clip once. If an experiment requires pruning some segments of the video clip, we simply remove the trajectories that are associated with the frames inside the segments.

\subsubsection{Fisher Vector encoding} \label{subsec:FV}

To encode features, we use Fisher Vector~\cite{Perronnin-et-al-ECCV10}. A Fisher Vector encodes both first and second order statistics between the feature descriptors and a Gaussian Mixture Model (GMM). In~\cite{Wang-Schmid-ICCV13}, Fisher Vector shows an improved performance over bag of features for action classification. Following \cite{Perronnin-et-al-ECCV10,Wang-Schmid-ICCV13}, we first reduce the dimension of DTDs by a factor of two using Principal Component Analysis (PCA). We set the number of Gaussians to $k = 256$ and randomly sample a subset of 1,000,000 features from the training sets to learn the GMM. There is one GMM for each feature type. A video sequence is represented by a $2dk$ dimensional Fisher Vector for each descriptor type, where $d$ is the descriptor dimension after performing PCA. As in \cite{Perronnin-et-al-ECCV10,Wang-Schmid-ICCV13}, we apply power ($\alpha = 0.5$) and $L_2$ normalization to the Fisher Vectors. We combine all descriptor types by concatenating their normalized Fisher Vectors, leading to a single feature vector of $109,056$ dimensions.

\subsubsection{Least-Squares SVM} \label{sec:LSSVM}
For recognition, we use Least-Squares Support Vector Machines (LSSVM) \cite{Suykens-Vamdewalle-NPL99}. LSSVM, also known as kernel Ridge regression \cite{Saunders-et-al-ICML98}, has been shown to perform equally well as SVM in many classification benchmarks~\cite{Suykens-et-al-02,Hoai-BMVC14,Vicente-et-al-ICCV15,Vicente-et-al-CVPR16}. LSSVM has a closed-form solution, which is a computational advantage over SVM~\cite{Vapnik-98}. We train 13 binary LSSVM classifiers for 13 action classes. For each action class, we train a one-versus-rest classifier where positive training examples are action samples from the class in consideration and negative training examples are action samples from other classes. 

\subsection{Results}

Table~\ref{tab:pruningBenefit} shows the Average Precision (AP) of the aforementioned action recognition method with and without pruning non-action shots. Here we measure performance using Average Precision, which is an accepted standard for action recognition (e.g.,~\cite{Marszalek-et-al-CVPR09,Patron-Perez-et-al-PAMI12,Hoai-et-al-BMVC14}). As can be seen, the ability to prune non-action shots leads to significant performance gain, which is shown in the last column of Table~\ref{tab:pruningBenefit}. The average performance gain is 13.7\%, and it is as high as 34.1\% for DriveCar.

\begin{figure}[t]
\begin{center}
\includegraphics[width=0.95\linewidth ]{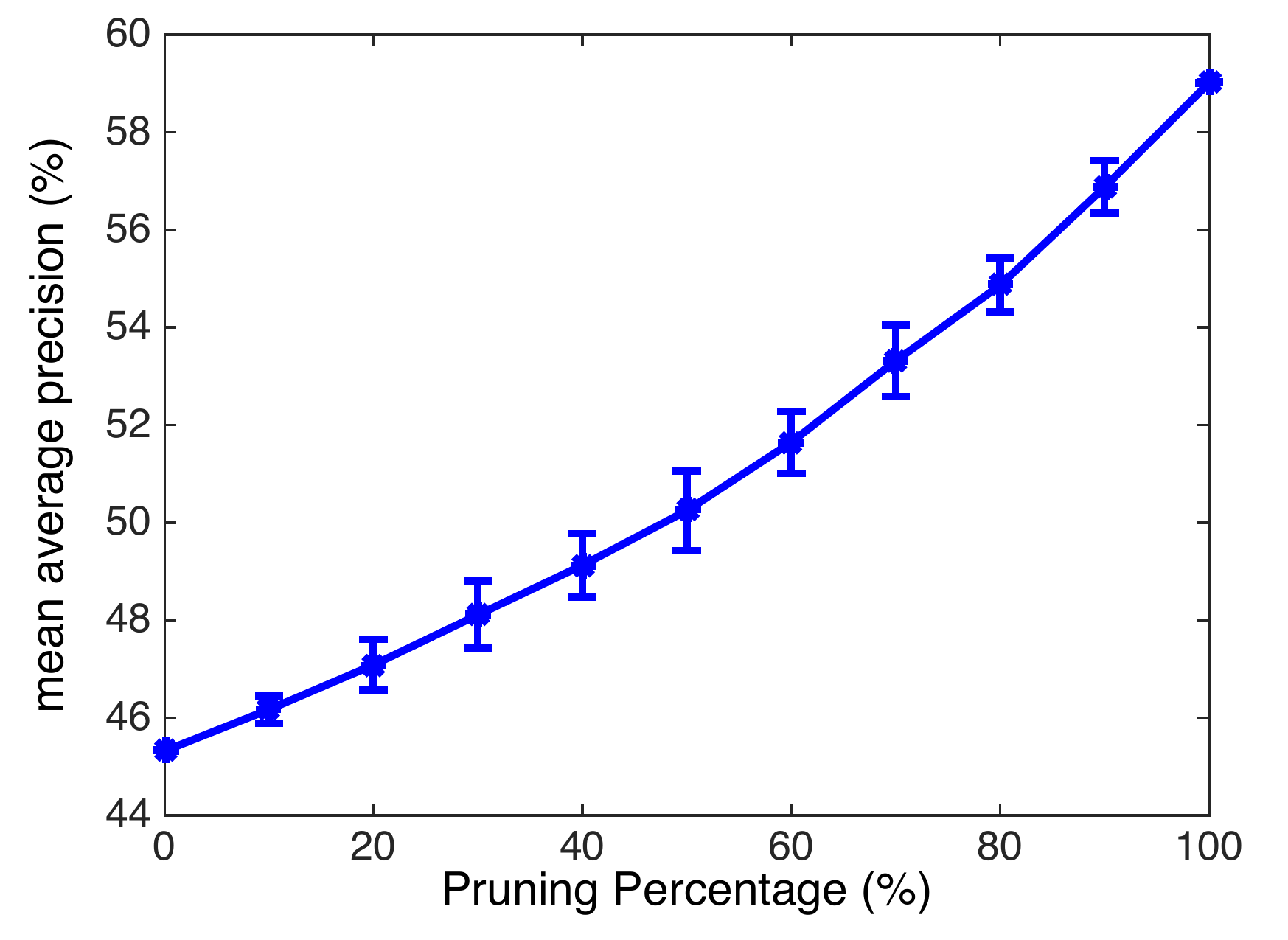}
\end{center}
\vskip -0.2in
   \caption{Mean average precision as a function of irrelevant-shot-pruning percentage. Even when the pruning percentage is far from 100\%, there is still significant performance gain.}
\label{fig:pruning_percentage}
\end{figure}

Table~\ref{tab:pruningBenefit} shows the benefits of an ideal situation where we can identify all non-action shots. This is perhaps unrealistic in practice, but we can still expect some improvement in the recognition performance even when we cannot remove all irrelevant shots. Figure~\ref{fig:pruning_percentage} shows the mean average precision for recognizing 13 action classes when the percentage for removing non-action shots is varied from 0 to 100\%. Given a percentage $p$ ($0\%\le p \le 100\%$), we randomly remove a non-action shot from the video with probability $p$. We repeat this experiment 20 times and compute the mean and standard deviation for the mean average precision. As can be observed, the performance gain increases as the removal percentage grows. The performance gain is significant even when we can only eliminate 40\% of the non-action shots.

\section{Non-action Classification} \label{sec:nonact_clf}

Having confirmed that removing non-action shots leads to large performance gain in the action recognition task, we describe in this section our approach for learning a classifier to differentiate between action shots from non-action shots. 

\subsection{Feature computation}

As also mentioned in the introduction, many non-action shots can be identified based on the size of the human characters, the amount of the motion, and the context of the shot in a longer video sequence (e.g., part of a dialog). To capture the discriminative information for classification, we propose to combine DTDs~\cite{Wang-Schmid-ICCV13} and deep-learned features from a Two-stream ConvNet~\cite{Simonyan-Zisserman-NIPS14}. These features lead to the state-of-the-art performance in many datasets~\cite{Marszalek-et-al-CVPR09,Kuehne-et-al-ICCV11,Soomro-et-al-TR12}. Recent experiments~\cite{wang2015action} have also suggested that they are complimentary to each other.

\subsubsection{Dense Trajectory Features} 

Dense Trajectory Features are extracted and used with Fisher Vector encoding as described in Sections~\ref{subsec:DTD} and~\ref{subsec:FV}. Note that each trajectory has a temporal span of 15 frames, and we assign each trajectory to the middle frame (the $8^{th}$ frame). Each frame is therefore associated with a set of trajectories. We compute an unnormalized Fisher Vector for each frame, and we have a sequence of frame-wise Fisher Vectors. Since unnormalized Fisher Vector is additive, the unnormalized Fisher Vector for a set of frames is the sum of  unnormalized Fisher Vectors for individual frames. Thus, given a sequence of frame-wise unnormalized Fisher Vectors for a video clip, we can efficiently compute the unnormalized Fisher Vector for any subsequence of the video clip. Finally the unnormalized Fisher Vector can be normalized to obtain the DTD feature representation for the subsequence.

\subsubsection{Deep-learning features}

We use deep-learning features from a Two-stream ConvNet~\cite{Simonyan-Zisserman-NIPS14}, which was proposed recently for human action recognition. In this paper, we use the pre-trained Two-Stream ConvNet provided by \citet{wang2015action} as a generic feature extractor. The model is trained on Split1 of UCF-101 dataset~\cite{Soomro-et-al-TR12}. This model contains  both a spatial and a temporal ConvNet~\cite{LeCun-et-al-NC89}. The spatial ConvNet is based on VGG-M-2048 model~\cite{Chatfield-et-al-BMVC14} and fine-tuned with image frames from videos; the temporal ConvNet have a similar structure, but its input is a set of 10 consecutive optical flow maps. 

Video frames are extracted at 25fps and subsequently resized to 256x340 pixels, having the aspect ratio of 4:3. Given the sequence of frames, we calculate the dense optical flow map between pairs of consecutive frames. We use the GPU version of TVL1 algorithm ~\cite{zach2007duality}, for its efficiency and accuracy. Following ~\cite{Simonyan-Zisserman-NIPS14}, we rescale and discretize the optical flow values into the integer range [0, 255], and store as image using JPEG compression. This greatly reduces the storage size.

\begin{figure}[t]
\begin{center}
\includegraphics[width=0.95\linewidth]{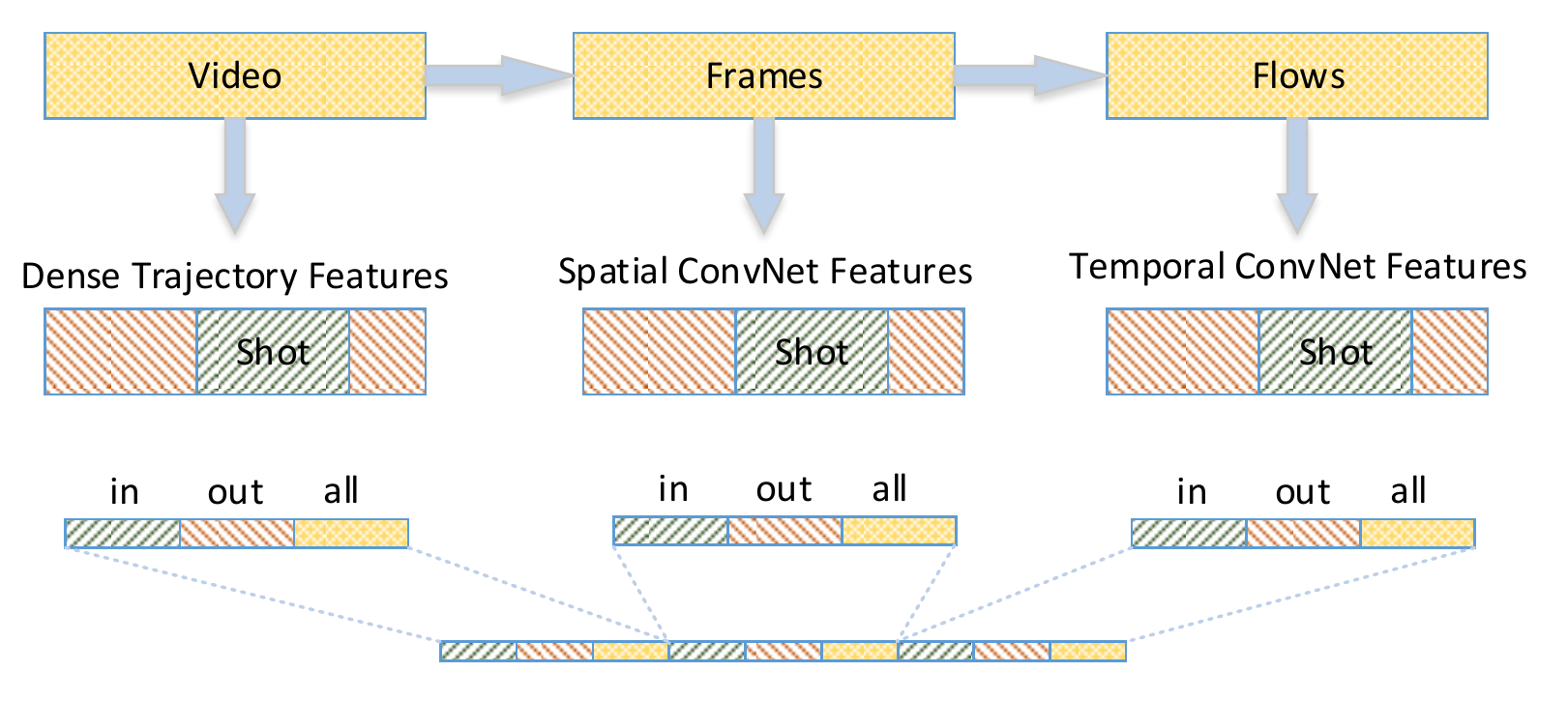}
\end{center}
\vskip -0.1in
   \caption{Features for non-action classification.}
\label{fig:clf_feature}
\end{figure}

\textbf{Spatial features.} The spatial ConvNet is based on the VGG-M-2048 model~\cite{Chatfield-et-al-BMVC14}. 
It requires input as an image region of size $224 \times 224 \times 3$ and outputs a 4096-dim feature vector at the FC6 layer. Note in VGG-M-2048, FC6 is a fully connected layer with 4096 dimensions, as opposed to FC7 which has 2048 dimensions.  To compute the feature vector for a video frame, we extract FC6 feature vectors at the center area and four corners of the frame as well as their left-right flipped images. Thus we obtain 10 feature vectors and average them to get a single 4096-dim spatial feature for a frame. Because consecutive frames are often similar, we only compute the spatial feature vector at every five frames. The feature vector for a set of frames (either from a contiguous video sequence or from the union of multiple disjoint video sequences) is the average of the feature vectors computed for the individual frames in the set.

\textbf{Temporal features.} The computation of temporal feature vectors is similar to that of spatial feature vectors. The only difference is that  the input to the temporal ConvNet must be a set of 10 consecutive optical flow maps. So, to compute the temporal feature vector for a frame $t$, we use the 10-frame volume that is centered at the frame $t$.

\subsubsection{Feature representation for a video shot} \label{subsec:ShotFT}

Consider a video shot $[s,e]$ which is a part of a longer video clip $[1, n]$. We first compute $f_{in}$ the feature vector for the video shot by concatenating the DTD feature vector, the spatial ConvNet feature vector, and the temporal ConvNet feature vector. Note that the DTD feature vector is power and $L_2$-normalized, while the spatial and temporal ConvNet feature vectors are $L_2$-normalized. In addition to $f_{in}$, we also compute two feature vectors $f_{out}$ and $f_{all}$ for the video frames outside $[s,e]$ and all the video frames, respectively. The ultimate feature vector to represent a video shot is taken as $[f_{in}, f_{out}, f_{all}]$, as illustrated in Figure~\ref{fig:clf_feature}. This feature vector encodes the appearance and motion of the video shot as well as its relative comparison with other video shots in its surrounding context.

\subsection{Training a non-action classifier}

We obtain a non-action classifier by training a Least-Squares SVM (Section~\ref{sec:LSSVM}) using data from the ActionThread dataset. This dataset is divided into disjoint train and test subsets, which contain 9724 and 9893 shots respectively. Within each subset, around 60\% of the shots are non-action. The feature representation for each shot combines both DTD and Two-stream ConvNet, as described in the previous section.

\subsection{Experiments and results}

We measure the performance of non-action classification on the test set of the ActionThread dataset. The test set contains 1,514 videos, with 5877 non-action shots and 4016 action shots. Table~\ref{tab:nonact_eval} shows the Average Precision of the non-action classifier based on different features. DTD outperforms the Spatial and Temporal features of the Two-Stream ConvNet when they are used individually. When combined, the Spatial and Temporal ConvNets achieve comparable result to DTD. The best performance is achieved when all feature types are combined. From now on, we will use the combined feature vector in all of our experiments.

The non-action classifier using the combined feature vectors achieve the average precision of 86.1\%. This classifier can be used to remove non-action shots and increase the signal-to-noise ratio of the action content in a video clip. In some cases, to minimize the chance of removing action shots, one might want to limit the number of shots removed for each video clip. Table~\ref{tab:ap_kshots} reports the Average Precision of the non-action classifier when the number of removed shots per video is limited to $k$, with $k=1,2,3,4$. As can be seen, limiting the number of removed shots per video can improve the average precision.

\setlength{\tabcolsep}{5pt}
\begin{table}[t]
\begin{center}
\begin{tabular}{lc}
\toprule
Feature  & AP \\
\midrule
%Chance                  & 60.0\% \\
%HoG                      & 76.1\% \\
%Motion Magnitude         & 76.7\% \\
Spatial (ConvNet)        &  80.8\%    \\
Temporal (ConvNet)       &  81.4\%    \\
Spatial+Temporal     &  84.1\%   \\
DTD       &  84.7\%   \\
DTD+Spatial+Temporal~~~~~~~~~~~~~  &\textbf{86.1\%}  \\
\bottomrule
\end{tabular}
\end{center}
\vskip -0.1in
\caption{\bf Evaluation of non-action shot classification.}
\label{tab:nonact_eval}
\end{table}

\setlength{\tabcolsep}{4pt}
\begin{table}[t]
\begin{center}
\begin{tabular}{ccccc}
\toprule
AP@k=1 & AP@k=2 & AP@k=3 & AP@k=4 & AP@k=$\infty$ \\
\midrule
92.6\% &   91.8\%  &  90.9\%  &  89.6\% & 86.1\%\\
\bottomrule
\end{tabular}
\end{center}
\vskip -0.1in
\caption{ Average Precision of the non-action classifier when we remove at most $k$ shots per video. $k=\infty$ corresponds to no constraint on the number of removed shots per video. \label{tab:ap_kshots}}
\end{table}

%\green{\subsubsection{Qualitative Results}}

\begin{figure}[h]
\begin{center}
\includegraphics[width=1\linewidth]{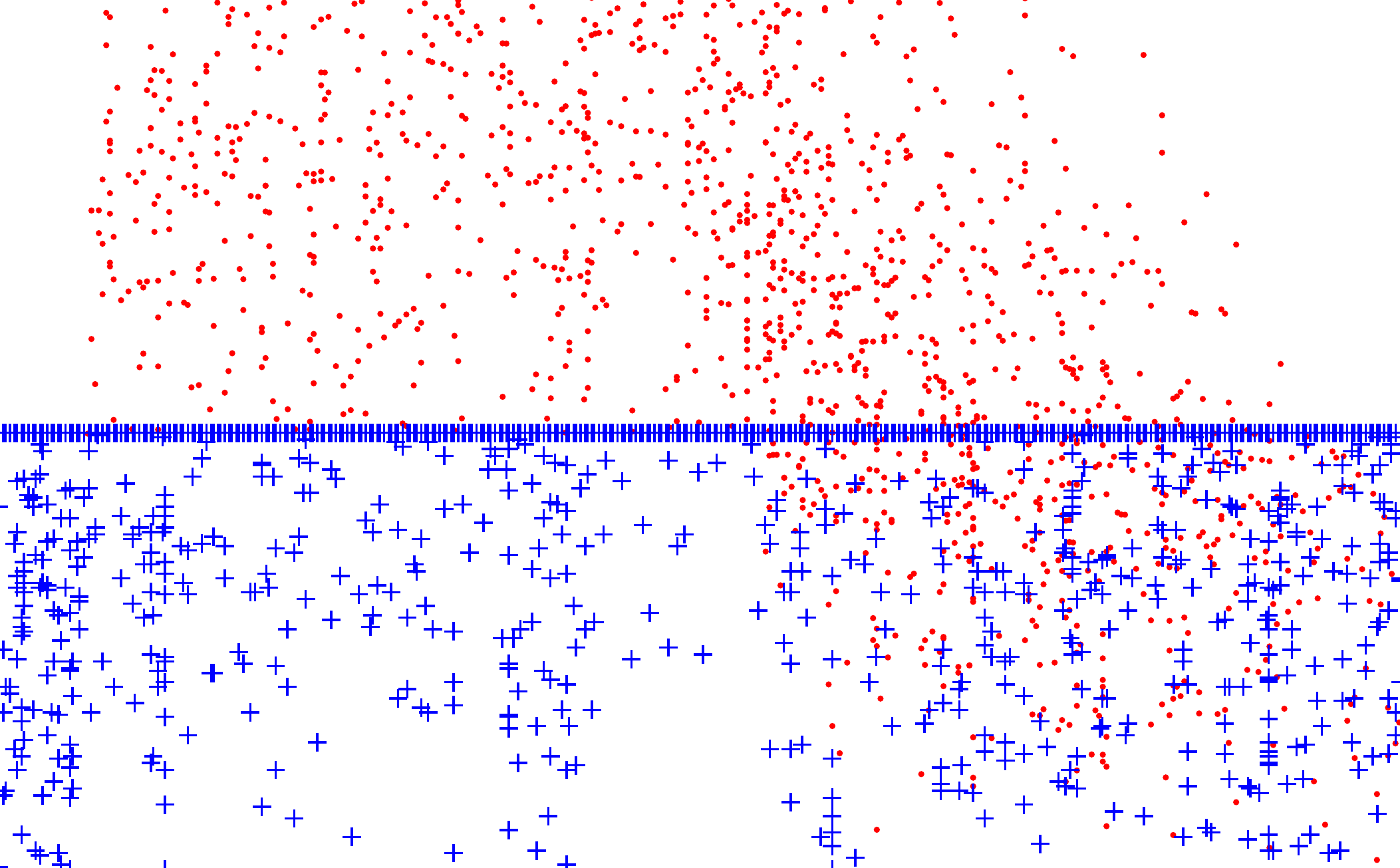}
\end{center}
\vskip -0.1in
\caption{{\bf Distribution of non-action scores.} Each column represents individual test video; red dots and blue pluses correspond to non-action and action shots, respectively. Shots shown above have higher confidence score than those below. Videos are ordered based on their classification AP. We uniformly sample and display 20\% of the test videos.}
\label{fig:score_dist}
\end{figure}

Figure \ref{fig:score_dist} shows the distribution of non-action confidence scores on video shots of the test set. Each column represents an individual video; the red dots and blue pluses on that column respectively correspond to the non-action and action shots in the video. The points above have higher confidence scores than those below. For visualization, we align each video with the horizontal bar in the middle using the maximum score of the action shots. As can be seen, action and non-action shots are separated fairly well.

\begin{figure}[h]
\begin{center}
\includegraphics[width=0.95\linewidth]{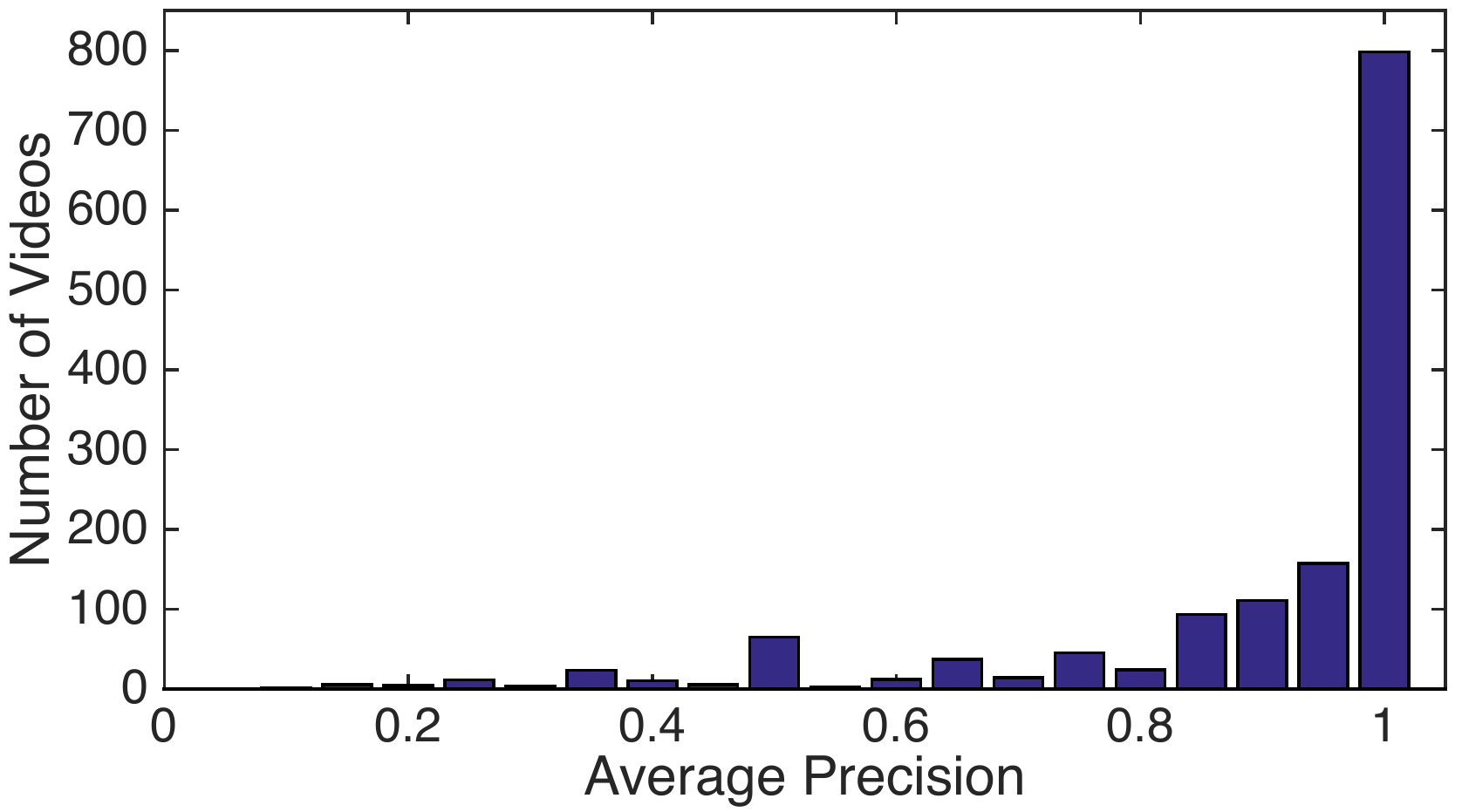}
\end{center}
\vskip -0.2in
   \caption{Distribution of the average precision  measured on individual test video clips for non-action classification. The majority of video clips have the average precision of 1, corresponding to perfect separation between action and non-action shots.}
\label{fig:ap_dist}
\end{figure}

We also examine the average precision of the non-action classifier for individual videos. Figure \ref{fig:ap_dist}  show the distribution of the average precisions computed for  individual videos. As can be seen, the big proportion of the videos have the average precision of 1, which means a perfect separation between action and non-action shots.

\subsubsection{Leave-One-Action-Out Generalizability} \label{subsec:loo}

\begin{figure}[t]
\begin{center}
\includegraphics[width=0.95\linewidth]{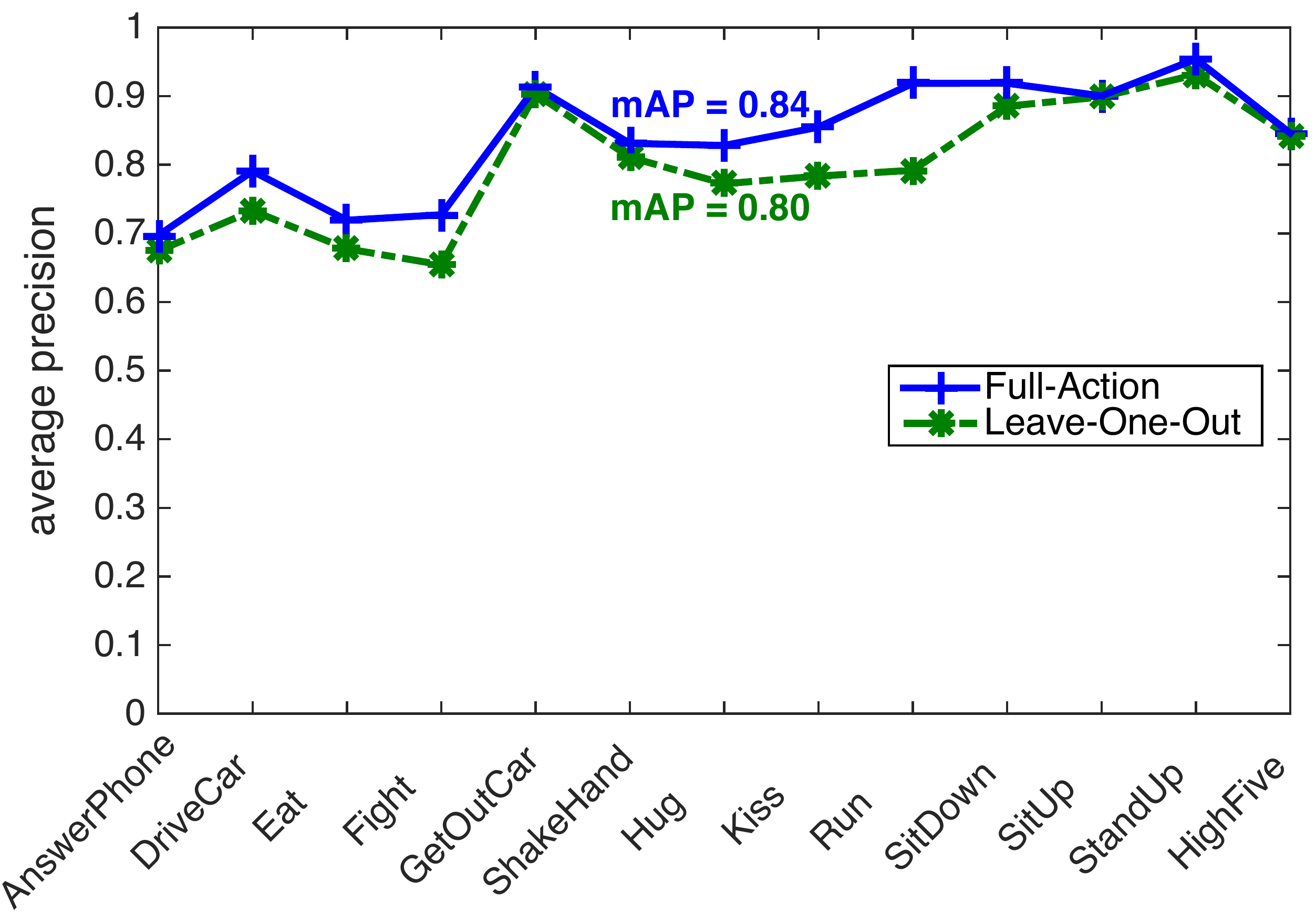}
\end{center}
\vskip -0.2in
   \caption{Comparison between the non-action classifier trained with all data (Full-Action) and the leave-one-class-out non-action classifiers (Leave-One-Out). The leave-one-out classifiers are comparable to the classifier trained on the full dataset. This demonstrates the ability to apply the non-action classifier to purify videos of unseen actions. \label{fig:loo_vs_full}}
\end{figure}

So far, we train a non-action classifier based on non-action shots from videos of several human actions. The classifier achieves the AP of 86.1\%, which is encouraging. However, it is still unclear whether this classifier can be used to identify non-action shots in videos of an action that is not among the set of actions in the training data. To test this, we consider the performance of the leave-one-class-out classifiers. In particular, we consider 13 action classes of the ActionThread dataset in turn. For each action class, we train a non-action classifier on a reduced training dataset where the videos of the action class in consideration are removed. The obtained non-action classifier is used to identify the non-action shots in the videos of the left-out action class. We compare the average precision of this classifier and the classifier trained with all data. The results are shown in Figure~\ref{fig:loo_vs_full}. As can be seen, the leave-one-class-out classifiers are comparable to the non-action classifier trained on the full dataset. This demonstrates the ability to apply the non-action classifier to purify videos of unseen actions.

\section{Non-action Classifier for Action Recognition} \label{sec:act_clf}
This section describes the benefits for using the non-action classifier for human action recognition. We also demonstrate the advantages of the generic non-action classifier over a set of action-specific non-action classifiers.

\subsection{Action recognition with non-action classifier} \label{subsec:act_rec_with_nonact_clf}
Our proposed algorithm is based on the action recognition system described in Section~\ref{subsec:rec_system}, using dense trajectory descriptors, Fisher Vector encoding, and LSSVM. The key difference is the incorporation of the non-action classifier to down-weight irrelevant non-action video segments.

Suppose a video is represented by a set of segments $\{W_i\}$ (discussed below). We first compute the normalized Fisher Vectors $\{\phi_i\}$ and non-action confidence scores $\{s_i\}$ of the segments. The feature vector for the video is taken as: 
\begin{equation}
\begin{split}
\phi  =\sum_{i}{w_i\phi_i}, \textrm{with }
w_i = \frac{e^{-\alpha s_i}}{\sum_{j}{e^{-\alpha s_j}}}
\end{split}
\label{equ:pooling}
\end{equation}
Here, we use $s_i$ to weight the contribution of $\phi_i$ using the softmax function. The higher $s_i$ the lower the weight is. 
The parameter $\alpha$ controls the balance between average pooling and max pooling. When $\alpha$ is 0, all $w_i$'s are the same, and this becomes average pooling. If $\alpha = \infty$, only one segment has the weight of 1, while the weights of others are 0. This is equivalent to max pooling. By tuning $\alpha$, we can have a good balance between average pooling and max pooling. Here we propose to weight the contribution of video segments instead of removing non-action segments because the non-action classifier is imperfect (even though the AP is as high as 92.6\% if we only remove the most confidence segment in each video). 

Our approach for weighting the segment contribution is not limited to any segment generation scheme. Any segment proposal method, including random selection, shot-based division, or an action proposal method (if exists), can be incorporated in our framework. In this paper, we propose to use video segments that are generated by using a 25-frame sliding window. Because videos are at 25fps, each segment corresponds to one second of a video. One second is neither too short nor too long, and we can use the non-action classifier to determine its non-action score. Empirically, the action recognition performance is not sensitive to the segment length around one second. 
Notably, the segments are not taken as video shots for two reasons. First, the shot classifier itself is imperfect. Second, even for a shot that is not considered as non-action, it might be long and contains many parts that do not portray human action.

%\textbf{Evaluation.} 
To verify the advantage of our action-weighted feature representation, we train 13 action classifiers using Least-Squares SVM with linear kernel as described in Section~\ref{sec:LSSVM}. Subsequently, we use another softmax function to normalize scores of different action classes. 
Note that we apply the segment generation and weighting on both training and test videos. For each video, we compute a single Fisher Vector as described in Equation (\ref{equ:pooling}). 
We report the average precision in the $4^{th}$ column of Table~\ref{tab:pooling_result}. This method achieves the mean average precision of 52.1\%, which is significantly higher than 45.3\%, the recognition performance without using the non-action classifier. The improvement is 6.8\%, which is comparable to the ability to remove 65\% of non-action shots as shown in  Figure \ref{fig:pruning_percentage}. 

Since the Two-stream ConvNet features are already extracted for computing the non-action score, we can also fuse them into the segment representation $\{\phi_i\}$. As shown in the last three columns of Table~\ref{tab:pooling_result}, our method as well as the baselines can benefit from the addition of CNN features.

\begin{table}[t]
\begin{center}
\begin{tabular}{lccc|ccc}
\toprule
\small{Feature} & \multicolumn{3}{c|}{DTD} & \multicolumn{3}{c}{DTD + CNN}\\
\midrule 
\small{Pruning} & \footnotesize{None} & \footnotesize{Specific} & \footnotesize{Generic} & \footnotesize{None} & \footnotesize{Specific} & \footnotesize{Generic} \\
\midrule
\small{Ans.Phone}   & 29.2  & 32.0 & \textbf{34.7} & 33.0 & 37.8 & \textbf{39.1} \\
\small{DriveCar}    & 41.4  & 53.1 & \textbf{55.6} & 45.9 & \textbf{56.9} & 55.9 \\
\small{Eat}         & 33.2  & 36.6 & \textbf{42.3} & 37.9 & 41.8 & \textbf{47.2} \\
\small{Fight}       & 63.0  & 63.2 & \textbf{68.1} & 63.0 & 62.7 & \textbf{68.9} \\
\small{GetOutCar}   & 23.8  & 30.9 & \textbf{30.9} & 28.0 & 35.1 & \textbf{38.6} \\
\small{ShakeHand}   & 42.9  & 51.5 & \textbf{58.3} & 44.5 & 53.3 & \textbf{60.9} \\
\small{Hug}         & 50.1  & 53.4 & \textbf{56.1} & 53.0 & 57.2 & \textbf{59.0} \\
\small{Kiss}        & 65.0  & \textbf{70.2} & 68.2 & 67.9 & \textbf{73.5} & 70.9 \\
\small{Run}         & 85.2  & 89.3 & \textbf{89.9} & 87.3 & 90.7 & \textbf{91.1} \\
\small{SitDown}     & 60.9  & 62.8 & \textbf{65.6} & 60.6 & 64.1 & \textbf{66.5} \\
\small{SitUp}       &  9.5  &  9.0 & \textbf{12.1} & 10.1 & 10.3 & \textbf{13.8} \\
\small{StandUp}     & 40.4  & 39.7 & \textbf{43.9} & 42.9 & 43.3 & \textbf{48.1} \\
\small{HighFive}    & 44.7  & 48.8 & \textbf{51.2} & 49.9 & 53.3 & \textbf{55.3} \\
\midrule
\small{Mean}        & 45.3  & 49.3 & \textbf{52.1} & 48.0 & 52.3 & \textbf{55.0} \\
\bottomrule
\end{tabular}
\end{center}
\vskip -0.1in
\caption{{\bf Action recognition performance with generic or specific non-action classifiers.} }
\label{tab:pooling_result}
\end{table}

\subsection{Action-specific non-action classifiers}
As an alternative to learning the generic non-action classifier, we can instead train for each action a specific non-action classifier. The pipeline of this approach is similar to the one described in last section with a few differences.  First, the determination of non-action shots is specific to an action. In training, we label a shot as non-action if it does not contain the action in consideration, regardless of the existence of other actions. Second, for representing a video, we use the action-specific non-action classifier and compute a different Fisher Vector feature for each action, as opposed to having the same feature vector for all actions. Thus we can train/test the classifier for each action based on the Fisher Vectors pruned by the corresponding specific non-action classifier. We report the average precision in the $3^{rd}$ column of Table~\ref{tab:pooling_result}. As can be seen, using the action-specific classifiers improves the recognition performance (compared with No Pruning in Table~\ref{tab:pooling_result}), but it is still outperformed by the method that uses the generic non-action classifier. This is probably because the generic classifier can be used to improve the action-to-noise ratio in videos of all action categories, while it is only meaningful to apply an action-specific classifier to videos of a specific category.

\subsection{VideoDarwin with non-action classifier} \label{sec:darwin}

To understand whether the benefits of using the non-action classifier is limited to the Fisher Vector encoding, we experiment with a method where we integrate the non-action classifier with VideoDarwin~\cite{fernando2015modeling}, a feature encoding method to capture the temporal evolution in a video sequence. VideoDarwin was proposed recently, and achieved the state-of-the-art performance on a number of datasets. It assumes every frame of a video clip carries some information about the action, and the total information about the action in a video segment correlates with the length. To capture this, VideoDarwin learns a Support Vector Regression (SVR) that maps the feature vector computed for each segment to its own length. More precisely, suppose $\phi_i$ is the feature vector for the $i^{th}$ frame. VideoDarwin learn the parameters $\u$ of an SVR such that: 
\begin{equation}
SVR(\u): \sum_{i=1}^{k}{\phi_i} \rightarrow \sum_{i=1}^{k}{1}, ~~k=1..N
\label{equ:darwin}
\end{equation}
The learned parameter vector $\u$ is then used as the feature representation for the video clip. Notably, this formulation is based on the Matlab implementation\footnote{https://bitbucket.org/bfernando/videodarwin} of the authors, and it is slightly different from the formulation given in the paper~\cite{fernando2015modeling} in which Ranking SVM is used instead. 

However, the assumption of VideoDarwin that every frame carries some information about the action does not hold due to the existence of non-action shots. To address this problem, we propose VideoDarwin++, a reformulated version that incorporating the outputs of a non-action classifier. VideoDarwin++ learns the parameters $\u$ of an SVR such that: 
\begin{equation}
SVR(\u): \sum_{i=1}^{k}{w_i \phi_i} \rightarrow \sum_{i=1}^{k}{w_i}, ~~k=1..N
\label{equ:newdarwin}
\end{equation}
Here the amount of information about an action in a segment does not solely depend on the length, but by the weights that are calculated based on the non-action scores.

Table~\ref{tab:darwin_result} compares the performance of VideoDarwin features, with and without the using non-action classifier. As can be seen, the non-action classifier provides benefits to this type of feature encoding.

\begin{table}[h]
\begin{center}
\begin{tabular}{lccc}
\toprule
 & ~~VideoDarwin & VideoDarwin++ \\
\midrule
Ans.Phone    & 32.3 & \textbf{37.6} \\
DriveCar     & 54.8 & \textbf{55.3} \\
Eat          & 34.8 & \textbf{42.5} \\
Fight        & 56.2 & \textbf{66.0} \\
GetOutCar    & 31.4 & \textbf{32.5} \\
ShakeHand    & 50.2 & \textbf{57.4} \\
Hug          & 51.7 & \textbf{56.5} \\
Kiss         & 65.2 & \textbf{69.0} \\
Run          & 84.7 & \textbf{88.9} \\
SitDown      & 62.7 & \textbf{66.2} \\
SitUp        & 10.5 & \textbf{12.9} \\
StandUp      & 40.1 & \textbf{45.3} \\
HighFive     & 41.5 & \textbf{47.9} \\
\midrule
Mean         & 47.4 & \textbf{52.2} \\
\bottomrule
\end{tabular}
\end{center}
\vskip -0.05in
\caption{{\bf Action recognition performance using VideoDarwin and VideoDarwin++ for feature encoding.} VideoDarwin++ is the reformulated version of VideoDarwin that incorporates the outputs of a non-action classifier.}
\label{tab:darwin_result}
\end{table}

\subsection{Cross-dataset generalization} \label{sec:transfer}

We have demonstrated the generalizability of the non-action classifier to videos in the test set and videos of left-out action categories in Section \ref{sec:nonact_clf}. Now we further study the benefits of the non-action classifier to action recognition task in completely different datasets. 

\subsubsection{Performance on Hollywood2}

We first consider Hollywood2 dataset~\cite{Marszalek-et-al-CVPR09}, which includes 12 actions and 1,707 videos collected from 69 Hollywood movies. 
We first divide videos into shots using a shot boundary detection algorithm based on HOG~\cite{Dalal-Triggs-CVPR05} and SIFT~\cite{Lowe-IJCV04} features, then manually label each shot for action occurrence or absence. Only 31\% are non-action shots, and this is `cleaner' than ActionThread. We apply the non-action classifier learned from ActionThread onto the Hollywood2 dataset and report in Table \ref{tab:hollywood2_result} the action recognition results with and without using the non-action classifier. 

\iffalse
\begin{table}[t]
\begin{center}
\begin{tabular}{lrrcc}
\toprule
        & \#Video & \#Shot & \#Non-Action Shot (\%) & AP \\
\midrule
Train   & 823 & 2394 & 760  ~(31.7\%) & 75.56\%\\
Test    & 884 & 2982 & 925  ~(31.0\%) & 70.80\%\\
\bottomrule
\end{tabular}
\vskip -0.1in
\end{center}
\caption{{\bf Video counts  and the percentage of non-action shots in the Hollywood2 dataset}. }
\label{tab:Hollywood2}
\end{table}
\fi

\begin{table}[t]
\begin{center}
\begin{tabular}{lcc|cc}
\toprule
Feature & \multicolumn{2}{c|}{DTD} & \multicolumn{2}{c}{DTD + CNN}\\
\midrule
Pruning & ~None~ & Generic & ~None~ & Generic \\
\midrule
Ans.Phone    & 29.3 & \textbf{33.3} & 36.0 & \textbf{39.9} \\
DriveCar     & 94.2 & \textbf{95.2} & 95.8 & \textbf{96.4} \\
Eat          & 64.2 & \textbf{66.4} & 63.9 & \textbf{68.4} \\
FightPerson  & 85.9 & \textbf{89.0} & 86.0 & \textbf{89.2} \\
GetOutCar    & 62.4 & \textbf{69.5} & 70.9 & \textbf{77.3} \\
HandShake    & 45.5 & \textbf{48.1} & 52.8 & \textbf{57.0} \\
HugPerson    & 50.0 & \textbf{51.5} & 51.7 & \textbf{55.1} \\
Kiss         & 64.0 & \textbf{64.0} & 67.9 & \textbf{68.5} \\
Run          & 86.4 & \textbf{89.5} & 89.1 & \textbf{92.5} \\
SitDown      & 77.4 & \textbf{82.6} & 77.9 & \textbf{82.3} \\
SitUp        & 32.3 & \textbf{39.2} & 33.4 & \textbf{41.8} \\
StandUp      & 78.9 & \textbf{81.4} & 80.5 & \textbf{82.9} \\
\midrule
Mean         & 64.2 & \textbf{67.5} & 67.2 & \textbf{71.0} \\
\bottomrule
\end{tabular}
\vskip -0.1in
\end{center}
\caption{{\bf Action recognition performance on Hollywood2.}}
\label{tab:hollywood2_result}
\end{table}

\subsubsection{Recognizing unseen action categories. }

We also collected human action samples of six actions that are not contained in the ActionThread dataset. Similar to the collection of ActionThread and Hollywood2, we extracted 100 video clips for each action using script mining, and manually examined and accepted those with the human action inside. Finally we have a dataset of 339 videos for 6 actions. We randomly split them into a training set (170 videos) and a test set (169 videos). Table~\ref{tab:otheraction_result} shows the recognition performance with and without using the non-action classifier (trained on the ActionThread dataset). As can be seen, the benefits of the non-action classifier can be generalized to action categories that do not exist in the training set of the non-action classifier. 

\begin{table}[t]
\begin{center}
\begin{tabular}{lcc|cc}
\toprule
Feature & \multicolumn{2}{c|}{DTD} & \multicolumn{2}{c}{DTD + CNN}\\
\midrule
Pruning & ~None~ & Generic & ~None~ & Generic \\
\midrule
CloseDoor       & 39.1 & \textbf{40.6} & 38.2 & \textbf{39.4} \\
OpenDoor        & 63.0 & \textbf{66.0} & 64.3 & \textbf{66.6} \\
Downstairs      & 89.2 & \textbf{93.0} & 89.6 & \textbf{92.9} \\
Dance           & 67.4 & \textbf{74.4} & 67.4 & \textbf{75.3} \\
Drink           & \textbf{68.9} & 68.6 & 67.9 & \textbf{68.7} \\
Applause        & 71.5 & \textbf{81.5} & 73.8 & \textbf{81.8} \\
\midrule
Mean            & 66.5 & \textbf{70.7} & 66.9 & \textbf{70.8} \\
\bottomrule
\end{tabular}
\end{center}
\vskip -0.1in
\caption{{\bf Action recognition performance on 6 unseen actions.}}
\vskip -0.1in
\label{tab:otheraction_result}
\end{table}

\section{Conclusions} \label{sec:conclusion}
\vskip -0.1in
We have studied the benefits of removing non-action shots and proposed a method for detecting them. Our detector is based on Dense Trajectories Descriptors and Two-stream ConvNet features. This  detector achieves an average precision of 86\%, and it can be used to down-weight the contribution of irrelevant segments in the computation of a feature vector to represent a video clip. This approach significantly improves the performance of a recognition system. In our experiments, the improvement is equivalent to the ability to remove 65\% of non-action shots without any false positive. Although the non-action classifier is far from perfect, it makes a good step towards the ultimate solution for human action recognition in realistic video. 

\noindent{\bf Acknowledgment.} This project is supported by the National Science Foundation Award IIS-1566248.

{\small
\bibliographystyle{abbrvnat}
\bibliography{longstrings,pubs,egbib}
}

\end{document}

%% file: definitions.tex
\def\mA{\mathcal{A}}
\def\mB{\mathcal{B}}
\def\mC{\mathcal{C}}
\def\mD{\mathcal{D}}
\def\mE{\mathcal{E}}
\def\mF{\mathcal{F}}
\def\mG{\mathcal{G}}
\def\mH{\mathcal{H}}
\def\mI{\mathcal{I}}
\def\mJ{\mathcal{J}}
\def\mK{\mathcal{K}}
\def\mL{\mathcal{L}}
\def\mM{\mathcal{M}}
\def\mN{\mathcal{N}}
\def\mO{\mathcal{O}}
\def\mP{\mathcal{P}}
\def\mQ{\mathcal{Q}}
\def\mR{\mathcal{R}}
\def\mS{\mathcal{S}}
\def\mT{\mathcal{T}}
\def\mU{\mathcal{U}}
\def\mV{\mathcal{V}}
\def\mW{\mathcal{W}}
\def\mX{\mathcal{X}}
\def\mY{\mathcal{Y}}
\def\mZ{\mathcal{Z}}

\def\1n{\mathbf{1}_n}
\def\0{\mathbf{0}}
\def\1{\mathbf{1}}

\def\A{{\bf A}}
\def\B{{\bf B}}
\def\C{{\bf C}}
\def\D{{\bf D}}
\def\E{{\bf E}}
\def\F{{\bf F}}
\def\G{{\bf G}}
\def\H{{\bf H}}
\def\I{{\bf I}}
\def\J{{\bf J}}
\def\K{{\bf K}}
\def\L{{\bf L}}
\def\M{{\bf M}}
\def\N{{\bf N}}
\def\O{{\bf O}}
\def\P{{\bf P}}
\def\Q{{\bf Q}}
\def\R{{\bf R}}
\def\S{{\bf S}}
\def\T{{\bf T}}
\def\U{{\bf U}}
\def\V{{\bf V}}
\def\W{{\bf W}}
\def\X{{\bf X}}
\def\Y{{\bf Y}}
\def\Z{{\bf Z}}

\def\a{{\bf a}}
\def\b{{\bf b}}
\def\c{{\bf c}}
\def\d{{\bf d}}
\def\e{{\bf e}}
\def\f{{\bf f}}
\def\g{{\bf g}}
\def\h{{\bf h}}
\def\i{{\bf i}}
\def\j{{\bf j}}
\def\k{{\bf k}}
\def\l{{\bf l}}
\def\m{{\bf m}}
\def\n{{\bf n}}
\def\o{{\bf o}}
\def\p{{\bf p}}
\def\q{{\bf q}}
\def\r{{\bf r}}
\def\s{{\bf s}}
\def\t{{\bf t}}
\def\u{{\bf u}}
\def\v{{\bf v}}
\def\w{{\bf w}}
\def\x{{\bf x}}
\def\y{{\bf y}}
\def\z{{\bf z}}

\def\balpha{\mbox{\boldmath{$\alpha$}}}
\def\bbeta{\mbox{\boldmath{$\beta$}}}
\def\bdelta{\mbox{\boldmath{$\delta$}}}
\def\bgamma{\mbox{\boldmath{$\gamma$}}}
\def\blambda{\mbox{\boldmath{$\lambda$}}}
\def\bsigma{\mbox{\boldmath{$\sigma$}}}
\def\btheta{\mbox{\boldmath{$\theta$}}}
\def\bomega{\mbox{\boldmath{$\omega$}}}
\def\bxi{\mbox{\boldmath{$\xi$}}}
\def\bnu{\mbox{\boldmath{$\nu$}}}                                  
\def\bphi{\mbox{\boldmath{$\phi$}}}

\def\bDelta{\mbox{\boldmath{$\Delta$}}}
\def\bOmega{\mbox{\boldmath{$\Omega$}}}
\def\bPhi{\mbox{\boldmath{$\Phi$}}}
\def\bLambda{\mbox{\boldmath{$\Lambda$}}}
\def\bSigma{\mbox{\boldmath{$\Sigma$}}}
\def\bGamma{\mbox{\boldmath{$\Gamma$}}}

\newcommand{\myminimum}[1]{\mathop{\textrm{minimum}}_{#1}}
\newcommand{\mymaximum}[1]{\mathop{\textrm{maximum}}_{#1}}    
\newcommand{\mymin}[1]{\mathop{\textrm{minimize}}_{#1}}
\newcommand{\mymax}[1]{\mathop{\textrm{maximize}}_{#1}}
\newcommand{\mymins}[1]{\mathop{\textrm{min.}}_{#1}}
\newcommand{\mymaxs}[1]{\mathop{\textrm{max.}}_{#1}}  
\newcommand{\myargmin}[1]{\mathop{\textrm{argmin}}_{#1}} 
\newcommand{\myargmax}[1]{\mathop{\textrm{argmax}}_{#1}} 
\newcommand{\myst}{\textrm{s.t. }}

\newcommand{\denselist}{\itemsep -1pt}
\newcommand{\sparselist}{\itemsep 1pt}

\definecolor{pink}{rgb}{0.9,0.5,0.5}
\definecolor{purple}{rgb}{0.5, 0.4, 0.8}   
\definecolor{gray}{rgb}{0.3, 0.3, 0.3}
\definecolor{mygreen}{rgb}{0.2, 0.6, 0.2}

\newcommand{\cyan}[1]{\textcolor{cyan}{#1}}
\newcommand{\red}[1]{\textcolor{red}{#1}}  
\newcommand{\blue}[1]{\textcolor{blue}{#1}}
\newcommand{\magenta}[1]{\textcolor{magenta}{#1}}
\newcommand{\pink}[1]{\textcolor{pink}{#1}}
\newcommand{\green}[1]{\textcolor{green}{#1}} 
\newcommand{\gray}[1]{\textcolor{gray}{#1}}    
\newcommand{\mygreen}[1]{\textcolor{mygreen}{#1}}    
\newcommand{\purple}[1]{\textcolor{purple}{#1}}       

\definecolor{greena}{rgb}{0.4, 0.5, 0.1}
\newcommand{\greena}[1]{\textcolor{greena}{#1}}

\definecolor{bluea}{rgb}{0, 0.4, 0.6}
\newcommand{\bluea}[1]{\textcolor{bluea}{#1}}
\definecolor{reda}{rgb}{0.6, 0.2, 0.1}
\newcommand{\reda}[1]{\textcolor{reda}{#1}}

\def\changemargin#1#2{\list{}{\rightmargin#2\leftmargin#1}\item[]}
\let\endchangemargin=\endlist
                                               
\newcommand{\cm}[1]{}

%% file: NAC.bbl
\begin{thebibliography}{40}
\providecommand{\natexlab}[1]{#1}
\providecommand{\url}[1]{\texttt{#1}}
\expandafter\ifx\csname urlstyle\endcsname\relax
  \providecommand{\doi}[1]{doi: #1}\else
  \providecommand{\doi}{doi: \begingroup \urlstyle{rm}\Url}\fi

\bibitem[Aggarwal and Ryoo(2011)]{Aggarwal-Ryoo-ACMCS11}
J.~Aggarwal and M.~Ryoo.
\newblock Human activity analysis: A review.
\newblock \emph{ACM Computing Surveys}, 43\penalty0 (3), 2011.

\bibitem[Alexe et~al.(2010)Alexe, Deselares, and Ferrari]{Alexe-et-al-CVPR10}
B.~Alexe, T.~Deselares, and V.~Ferrari.
\newblock What is an object?
\newblock In \emph{Proceedings of the IEEE Conference on Computer Vision and
  Pattern Recognition}, 2010.

\bibitem[Alexe et~al.(2012)Alexe, Deselaers, and Ferrari]{alexe2012measuring}
B.~Alexe, T.~Deselaers, and V.~Ferrari.
\newblock Measuring the objectness of image windows.
\newblock \emph{IEEE Transactions on Pattern Analysis and Machine
  Intelligence}, 34\penalty0 (11):\penalty0 2189--2202, 2012.

\bibitem[Chatfield et~al.(2014)Chatfield, Simonyan, Vedaldi, and
  Zisserman]{Chatfield-et-al-BMVC14}
K.~Chatfield, K.~Simonyan, A.~Vedaldi, and A.~Zisserman.
\newblock Return of the devil in the details: Delving deep into convolutional
  nets.
\newblock In \emph{Proceedings of the British Machine Vision Conference}, 2014.

\bibitem[Chen et~al.(2014)Chen, Xiong, Xu, and Corso]{chen2014actionness}
W.~Chen, C.~Xiong, R.~Xu, and J.~J. Corso.
\newblock Actionness ranking with lattice conditional ordinal random fields.
\newblock In \emph{Proceedings of the IEEE Conference on Computer Vision and
  Pattern Recognition}, 2014.

\bibitem[Dalal and Triggs(2005)]{Dalal-Triggs-CVPR05}
N.~Dalal and B.~Triggs.
\newblock Histograms of oriented gradients for human detection.
\newblock In \emph{Proceedings of the IEEE Conference on Computer Vision and
  Pattern Recognition}, 2005.

\bibitem[Feichtenhofer et~al.(2015)Feichtenhofer, Pinz, and
  Wildes]{feichtenhofer2015dynamically}
C.~Feichtenhofer, A.~Pinz, and R.~P. Wildes.
\newblock Dynamically encoded actions based on spacetime saliency.
\newblock In \emph{Proceedings of the IEEE Conference on Computer Vision and
  Pattern Recognition}, 2015.

\bibitem[Fernando et~al.(2015)Fernando, Gavves, Oramas, Ghodrati, and
  Tuytelaars]{fernando2015modeling}
B.~Fernando, E.~Gavves, J.~Oramas, A.~Ghodrati, and T.~Tuytelaars.
\newblock Modeling video evolution for action recognition.
\newblock In \emph{Proceedings of the IEEE Conference on Computer Vision and
  Pattern Recognition}, 2015.

\bibitem[Gkioxari and Malik(2014)]{gkioxari2014finding}
G.~Gkioxari and J.~Malik.
\newblock Finding action tubes.
\newblock \emph{arXiv preprint arXiv:1411.6031}, 2014.

\bibitem[Hadfield and Bowden(2013)]{Hadfield-Bowden-CVPR13}
S.~Hadfield and R.~Bowden.
\newblock Hollywood {3D}: Recognizing actions in {3D} natural scenes.
\newblock In \emph{Proceedings of the IEEE Conference on Computer Vision and
  Pattern Recognition}, 2013.

\bibitem[Hoai(2014)]{Hoai-BMVC14}
M.~Hoai.
\newblock Regularized max pooling for image categorization.
\newblock In \emph{Proceedings of the British Machine Vision Conference}, 2014.

\bibitem[Hoai and Zisserman(2014{\natexlab{a}})]{Hoai-Zisserman-ACCV14}
M.~Hoai and A.~Zisserman.
\newblock Thread-safe: Towards recognizing human actions across shot
  boundaries.
\newblock In \emph{Proceedings of the Asian Conference on Computer Vision},
  2014{\natexlab{a}}.

\bibitem[Hoai and Zisserman(2014{\natexlab{b}})]{Hoai-Zisserman-ACCV14b}
M.~Hoai and A.~Zisserman.
\newblock Improving human action recognition using score distribution and
  ranking.
\newblock In \emph{Proceedings of the Asian Conference on Computer Vision},
  2014{\natexlab{b}}.

\bibitem[Hoai and Zisserman(2014{\natexlab{c}})]{Hoai-Zisserman-CVPR14}
M.~Hoai and A.~Zisserman.
\newblock Talking heads: Detecting humans and recognizing their interactions.
\newblock In \emph{Proceedings of the IEEE Conference on Computer Vision and
  Pattern Recognition}, 2014{\natexlab{c}}.

\bibitem[Hoai et~al.(2011)Hoai, Lan, and {De la Torre}]{Hoai-et-al-CVPR11}
M.~Hoai, Z.-Z. Lan, and F.~{De la Torre}.
\newblock Joint segmentation and classification of human actions in video.
\newblock In \emph{Proceedings of the IEEE Conference on Computer Vision and
  Pattern Recognition}, 2011.

\bibitem[Hoai et~al.(2014)Hoai, Ladicky, and Zisserman]{Hoai-et-al-BMVC14}
M.~Hoai, L.~Ladicky, and A.~Zisserman.
\newblock Action recognition from weak alignment of body parts.
\newblock In \emph{Proceedings of the British Machine Vision Conference}, 2014.

\bibitem[Hou and Zhang(2007)]{hou2007saliency}
X.~Hou and L.~Zhang.
\newblock Saliency detection: A spectral residual approach.
\newblock In \emph{Proceedings of the IEEE Conference on Computer Vision and
  Pattern Recognition}. IEEE, 2007.

\bibitem[Kuehne et~al.(2011)Kuehne, Jhuang, Garrote, Poggio, and
  Serre]{Kuehne-et-al-ICCV11}
H.~Kuehne, H.~Jhuang, E.~Garrote, T.~Poggio, and T.~Serre.
\newblock {HMDB}: A large video database for human motion recognition.
\newblock In \emph{Proceedings of the International Conference on Computer
  Vision}, 2011.

\bibitem[Lai et~al.(2014)Lai, Yu, Chen, and Chang]{lai2014video}
K.-T. Lai, F.~X. Yu, M.-S. Chen, and S.-F. Chang.
\newblock Video event detection by inferring temporal instance labels.
\newblock In \emph{Proceedings of the IEEE Conference on Computer Vision and
  Pattern Recognition}, 2014.

\bibitem[LeCun et~al.(1989)LeCun, Boser, Denker, and
  Henderson]{LeCun-et-al-NC89}
Y.~LeCun, B.~Boser, J.~S. Denker, and D.~Henderson.
\newblock Backpropagation applied to handwritten zip code recognition.
\newblock \emph{Neural Computation}, 1\penalty0 (4):\penalty0 541--551, 1989.

\bibitem[Lowe(2004)]{Lowe-IJCV04}
D.~Lowe.
\newblock Distinctive image features from scale-invariant keypoints.
\newblock \emph{International Journal of Computer Vision}, 60\penalty0
  (2):\penalty0 91--110, 2004.

\bibitem[Marszalek et~al.(2009)Marszalek, Laptev, and
  Schmid]{Marszalek-et-al-CVPR09}
M.~Marszalek, I.~Laptev, and C.~Schmid.
\newblock Actions in context.
\newblock In \emph{Proceedings of the IEEE Conference on Computer Vision and
  Pattern Recognition}, 2009.

\bibitem[Nguyen et~al.(2015)Nguyen, Song, and Yan]{nguyen2015stap}
T.~V. Nguyen, Z.~Song, and S.~Yan.
\newblock Stap: Spatial-temporal attention-aware pooling for action
  recognition.
\newblock \emph{IEEE Transactions on Circuits and Systems for Video
  Technology}, 25\penalty0 (1):\penalty0 77--86, 2015.

\bibitem[Ni et~al.(2015)Ni, Moulin, Yang, and Yan]{ni2015motion}
B.~Ni, P.~Moulin, X.~Yang, and S.~Yan.
\newblock Motion part regularization: Improving action recognition via
  trajectory selection.
\newblock In \emph{Proceedings of the IEEE Conference on Computer Vision and
  Pattern Recognition}, 2015.

\bibitem[Patron-Perez et~al.()Patron-Perez, Marszalek, Zisserman, and
  Reid]{Patron-Perez2010}
A.~Patron-Perez, M.~Marszalek, A.~Zisserman, and I.~Reid.
\newblock In \emph{Proceedings of the British Machine Vision Conference}.

\bibitem[Patron-Perez et~al.(2012)Patron-Perez, Marszalek, Reid, and
  Zisserman]{Patron-Perez-et-al-PAMI12}
A.~Patron-Perez, M.~Marszalek, I.~Reid, and A.~Zisserman.
\newblock Structured learning of human interactions in {TV} shows.
\newblock \emph{IEEE Transactions on Pattern Analysis and Machine
  Intelligence}, 34\penalty0 (12):\penalty0 2441--2453, 2012.

\bibitem[Perronnin et~al.(2010)Perronnin, S\'anchez, and
  Mensink]{Perronnin-et-al-ECCV10}
F.~Perronnin, J.~S\'anchez, and T.~Mensink.
\newblock Improving the fisher kernel for large-scale image classification.
\newblock In \emph{Proceedings of the European Conference on Computer Vision},
  2010.

\bibitem[Saunders et~al.(1998)Saunders, Gammerman, and
  Vovk]{Saunders-et-al-ICML98}
C.~Saunders, A.~Gammerman, and V.~Vovk.
\newblock Ridge regression learning algorithm in dual variables.
\newblock In \emph{Proceedings of the International Conference on Machine
  Learning}, 1998.

\bibitem[Simonyan and Zisserman(2014)]{Simonyan-Zisserman-NIPS14}
K.~Simonyan and A.~Zisserman.
\newblock Two-stream convolutional networks for action recognition in videos.
\newblock In \emph{Advances in Neural Information Processing Systems}, 2014.

\bibitem[Soomro et~al.(2012)Soomro, Zamir, and Shah]{Soomro-et-al-TR12}
K.~Soomro, A.~R. Zamir, and M.~Shah.
\newblock {UCF101}: A dataset of 101 human action classes from videos in the
  wild.
\newblock Technical Report CRCV-TR-12-01, University of Central Florida, 2012.

\bibitem[Suykens and Vandewalle(1999)]{Suykens-Vamdewalle-NPL99}
J.~A.~K. Suykens and J.~Vandewalle.
\newblock Least squares support vector machine classifiers.
\newblock \emph{Neural Processing Letters}, 9\penalty0 (3):\penalty0 293--300,
  1999.

\bibitem[Suykens et~al.(2002)Suykens, Gestel, Brabanter, DeMoor, and
  Vandewalle]{Suykens-et-al-02}
J.~A.~K. Suykens, T.~V. Gestel, J.~D. Brabanter, B.~DeMoor, and J.~Vandewalle.
\newblock \emph{Least Squares Support Vector Machines}.
\newblock World Scientific, 2002.

\bibitem[Uijlings et~al.(2013)Uijlings, van~de Sande, Gevers, and
  Smeulders]{uijlings2013selective}
J.~R. Uijlings, K.~E. van~de Sande, T.~Gevers, and A.~W. Smeulders.
\newblock Selective search for object recognition.
\newblock \emph{International Journal of Computer Vision}, 104\penalty0
  (2):\penalty0 154--171, 2013.

\bibitem[Vapnik(1998)]{Vapnik-98}
V.~Vapnik.
\newblock \emph{Statistical Learning Theory}.
\newblock Wiley, New York, NY, 1998.

\bibitem[Vicente et~al.(2015)Vicente, Hoai, and Samaras]{Vicente-et-al-ICCV15}
T.~Y. Vicente, M.~Hoai, and D.~Samaras.
\newblock Leave-one-out kernel optimization for shadow detection.
\newblock In \emph{Proceedings of the International Conference on Computer
  Vision}, 2015.

\bibitem[Vicente et~al.(2016)Vicente, Hoai, and Samaras]{Vicente-et-al-CVPR16}
T.~Y. Vicente, M.~Hoai, and D.~Samaras.
\newblock Noisy label recovery for shadow detection in unfamiliar domains.
\newblock In \emph{Proceedings of the IEEE Conference on Computer Vision and
  Pattern Recognition}, 2016.

\bibitem[Wang and Schmid(2013)]{Wang-Schmid-ICCV13}
H.~Wang and C.~Schmid.
\newblock Action recognition with improved trajectories.
\newblock In \emph{Proceedings of the International Conference on Computer
  Vision}, 2013.

\bibitem[Wang et~al.(2015)Wang, Qiao, and Tang]{wang2015action}
L.~Wang, Y.~Qiao, and X.~Tang.
\newblock Action recognition with trajectory-pooled deep-convolutional
  descriptors.
\newblock \emph{arXiv preprint arXiv:1505.04868}, 2015.

\bibitem[Zach et~al.(2007)Zach, Pock, and Bischof]{zach2007duality}
C.~Zach, T.~Pock, and H.~Bischof.
\newblock A duality based approach for realtime optical flow.
\newblock \emph{Pattern Recognition}, pages 214--223, 2007.

\bibitem[Zitnick and Doll{\'a}r(2014)]{zitnick2014edge}
C.~L. Zitnick and P.~Doll{\'a}r.
\newblock Edge boxes: Locating object proposals from edges.
\newblock In \emph{Proceedings of the European Conference on Computer Vision}.
  2014.

\end{thebibliography}
